%% file: root.tex
\documentclass[letterpaper, 10 pt, conference]{ieeeconf}  % Comment this line out if you need a4paper

\IEEEoverridecommandlockouts                              % This command is only needed if 
\usepackage{xcolor}
\usepackage{textcomp}
\usepackage[pdftex]{graphicx}
\usepackage{adjustbox}

\usepackage{subcaption}

\usepackage{times}
\usepackage{color, colortbl, xcolor}
\usepackage{multicol}
\usepackage[bookmarks=true]{hyperref}

\usepackage{amsfonts}
\usepackage{algorithmic}
\usepackage[ruled,vlined,titlenotnumbered]{algorithm2e} 
\usepackage{amsmath}

\usepackage{dsfont}
\usepackage{multicol}
\usepackage[pdftex]{graphicx}   
\usepackage{mathtools}
\usepackage{accents}
\usepackage{wrapfig}
\usepackage{siunitx}
\let\labelindent\relax
\usepackage{enumitem}
\usepackage{balance}

\usepackage{url}

\graphicspath{{./figs/}}    
\DeclareGraphicsExtensions{.pdf,.png}  

\usepackage[textfont=md,font=footnotesize]{caption}

\input{notations}

\title{\LARGE \bf
Gait Switching and Enhanced Stabilization of Walking Robots with Deep Learning-based Reachability: A Case Study on Two-link Walker
}
\author{Xingpeng Xia$^{*,1}$, Jason J. Choi$^{*2}$, Ayush Agrawal$^{2}$, Koushil Sreenath$^{2}$, Claire J. Tomlin$^{2}$, and Somil Bansal$^{3}$% <-this % stops a space
\thanks{* The first two authors contributed equally. This work is supported in part by the NSF Grant CMMI-1944722, the NSF CAREER Program under award 2240163, the NASA ULI on Safe Aviation Autonomy, and the DARPA Assured Autonomy and Assured Neuro Symbolic Learning and Reasoning (ANSR) programs. The work of Jason J. Choi received the support of a fellowship from Kwanjeong Educational Foundation, Korea.}% <-this % stops a space
\thanks{$^{1}$Tsinghua University, Beijing, China. $^{2}$University of California, Berkeley, CA, US.
$^{3}$University of Southern California, Los Angeles, CA, US
        {\tt\small xiaxp19@mails.tsinghua.edu.cn, jason.choi@berkeley.edu, somilban@usc.edu}}%
}

\begin{document}
\newtheorem{proposition}{Proposition}
\newtheorem{definition}{Definition}
\newtheorem{problem}{Problem}
\newtheorem{corollary}{Corollary}
\newtheorem{lemma}{Lemma}
\newtheorem{theorem}{Theorem}
\newtheorem{remark}{Remark}
\newtheorem{assumption}{Assumption}

\maketitle
\thispagestyle{empty}
\pagestyle{empty}

%%%%%%%%%%%%%%%%%%%%%%%%%%%%%%%%%%%%%%%%%%%%%%%%%%%%%%%%%%%%%%%%%%%%%%%%%%%%%%%%
\begin{abstract}
Learning-based approaches have recently shown notable success in legged locomotion. However, these approaches often lack accountability, necessitating empirical tests to determine their effectiveness. In this work, we are interested in designing a learning-based locomotion controller whose stability can be examined and guaranteed. This can be achieved by verifying regions of attraction (RoAs) of legged robots to their stable walking gaits. This is a non-trivial problem for legged robots due to their hybrid dynamics. Although previous work has shown the utility of Hamilton-Jacobi (HJ) reachability to solve this problem, its practicality was limited by its poor scalability. The core contribution of our work is the employment of a deep learning-based HJ reachability solution to the hybrid legged robot dynamics, which overcomes the previous work's limitation. With the learned reachability solution, first, we can estimate a library of RoAs for various gaits. Second, we can design a one-step predictive controller that effectively stabilizes to an individual gait within the verified RoA. Finally, we can devise a strategy that switches gaits, in response to external perturbations, whose feasibility is guided by the RoA analysis. We demonstrate our method in a two-link walker simulation, whose mathematical model is well established. Our method achieves improved stability than previous model-based methods, while ensuring transparency that was not present in the existing learning-based approaches.
\end{abstract}

%%%%%%%%%%%%%%%%%%%%%%%%%%%%%%%%%%%%%%%%%%%%%%%%%%%%%%%%%%%%%%%%%%%%%%%%%%%%%%%%
% \vspace{-0.5em}

\section{Introduction}

\subsection{Motivation \& Key Ideas}

Locomotion is one of the fundamental modes of mobility for robots. The recent success of deep reinforcement learning (RL) in achieving robust, stable walking for legged robots \cite{li2021reinforcement, siekmann2021blind, lee2020learning} underscores the strength of learning-based policies. Despite this success, the mechanisms underlying the RL policies remain opaque. For example, the RL policy in \cite{li2021reinforcement} stretches the leg when the robot is pushed to keep its balance. However, this action, derived from a black-box neural network, leaves us unable to explain the underlying rationale. Generally, we seek more than just empirical observations to determine why and when a policy works.

To address this limitation, our study focuses on developing explainable learning-based policies for stable locomotion. Specifically, we seek to elucidate
\begin{enumerate}[leftmargin=1.25em, labelindent=\parindent, listparindent=\parindent, labelwidth=0pt]
    \item the region in which the robot state can be perturbed without compromising its ability to return to a stable gait,
    \item the rationale for selecting a feasible target gait among the library of candidate gaits,
    \item the way of evaluating the feasibility of gait transition.
\end{enumerate}
By designing a learning-based policy which accompanies answers to these questions, we can furnish learning-based locomotion with a layer of interpretability and assurance.

The central idea of our approach to tackle these questions is to resort to the region of attraction (RoA) concept. The RoA defines the state space region from which a system can stabilize to a desired attractor---in the case of legged robots, a periodic walking gait.
A policy informed by the RoAs can determine when stabilization is feasible and how. Although RoA analysis is a classic topic in controls, its application to legged robots has been hindered by the robots' complex hybrid dynamics, with a few notable exceptions \cite{MANCHESTER20115801, posa2015stability, choi2022}. The method in \cite{choi2022} stands out for recovering largest portion of the RoA through reachability analysis, yet it is significantly limited by the computational demands of solving Hamilton-Jacobi (HJ) partial differential equations (PDEs) numerically.

Our approach builds upon \cite{choi2022} but circumvents the computational hurdles of the numerical method by leveraging neural networks to approximate the HJ PDE solutions. Adapting the approach in \cite{bansal2020deepreach}, we tailor the learning of HJ PDE solutions to the hybrid dynamics and diverse gaits of legged robots. Our walking policy is designed around the learned HJ PDE solution. By ensuring that it stabilizes to a feasible gait based on the RoA evaluation, we not only significantly enhance the stabilization capabilities of the robot, but also provide a transparent rationale behind the controller's decisions.

\vspace{-0.25em}

\subsection{Contributions}

\vspace{-0.25em}

The main novelty of this work is the first-ever application of the deep learning-based reachability to hybrid system dynamics and locomotion control design. Its primary purpose is to provide users with an RoA estimate from which the robot can stabilize to a set of stable walking gaits. With the learned value function, the solution of the HJ PDE, we are able to estimate RoAs of individual gaits, and their union shapes the feasible space for stable walking.

% With the learned value function, the solution of the HJ PDE, we are able to verify the RoA of each gait. The union of the verified RoAs shapes the feasible space for stable walking. % From the states encapsulated in the RoA, it is always feasible to find a control that stabilizes the system to the target gait. Thus, the union of the verified RoAs shapes the feasible space for stable walking. % In contrast, at a state that belongs to none of the RoAs, it might not be feasible to stabilize the robot, no matter how good the control policy is.

Second, we design a control policy that stabilizes to an individual gait, based on the learned HJ PDE solution. 
Rather than resorting to the optimal reachability policy that minimizes the Hamiltonian of the learned value function \cite{choi2022}, we devise a one-step predictive control design. It seeks the best control in minimizing the value function at the next timestep, which mitigates the learning errors of the neural network more effectively and achieves enhanced stabilization.

Finally, we devise an effective gait switching strategy whose feasibility is provable by the RoA analysis. This gait transition is inevitable, especially when the robot is subjected to unmodeled perturbations that lead the system state to escape the RoA of the current gait. By evaluating RoAs, we can cleverly select which gait the robot switches to, and continue stable walking without falling. 

We demonstrate the aforementioned contributions in a simple two-link walker robot simulation. Although the dynamics of the robot is simplistic as a legged robot, its low dimensionality allows us to conduct a detailed analysis of the verified RoAs and access to its detailed mathematical model allows a fair comparison to existing model-based control methods. Our experiment reveals that 1) RoAs of the gaits can be estimated accurately with the proposed learning-based framework, 2) the designed controller achieves a significantly higher success rate than model-based stabilization controllers like the hybrid zero dynamics-based input-output (IO) linearization controller \cite{westervelt2018feedback}, and nonlinear model predictive controller (NMPC) \cite{negri2020nonlinear}, and 3) gait transitions can be conducted stably when the robot is pursing a sequence of varying gaits or when it is subjected to strong perturbations.

\vspace{-0.25em}
\subsection{Related work}
\vspace{-0.25em}

\subsubsection{Learning-based approach for locomotion}
The success recipe for deep RL-based locomotion includes many elements such as high-fidelity simulations \cite{hwangbo2019learning}, good composite reward design \cite{li2024reinforcement}, appropriate training schemes like domain randomization \cite{li2021reinforcement} or curriculum learning \cite{li2024reinforcement}, and a suitable model architecture \cite{radosavovic2024humanoid}. % Accurate and efficient simulation of contacts and robot dynamics are essential for the training environment \cite{hwangbo2019learning}. Composition of multiple reward terms with carefully tuned weights guides the RL to achieve robust stable locomotion \cite{li2024reinforcement}. Domain randomization \cite{tobin2017domain, li2021reinforcement} is another essential element, which exposes the policy to random variations of dynamics and environments, for obtaining robust policies. Finally, the learning models should also be carefully selected; various design choices have to be made such as the used model class \cite{radosavovic2024humanoid}. 
All in all, carefully trained RL policies achieved state-of-the-art performance in locomotion. % Nevertheless, they lack explainability and still remain as a black-box component of the system.

\subsubsection{Model-based analysis and control design for locomotion}
Numerous mathematical tools are proposed for the analysis of stability of locomotion---Lyapunov methods \cite{MANCHESTER20115801, posa2015stability}, Poincar\'e map \cite{Motahar2016}, capturability \cite{koolen2012capturability}, contraction analysis \cite{burden2022infinitesimal}, Riemannian partition \cite{gu2022reactive}, and reachability \cite{choi2022, piovan2015reachability, ding2021hybrid}. Although each method has its unique strengths and drawbacks, verifying an invariant and stable domain is the central theme of most methods. 

The control design approaches also vary from IO linearizaiton \cite{sreenath2011compliant} to NMPC \cite{negri2020nonlinear, dosunmu2024demonstrating, choe2023seamless, li2023unified} and many others. The common challenge in these designs is addressing the varying gait sequence and the associated contact dynamics. The main benefit of our approach is that the hybrid dynamics are already accounted for in the construction of the value function, and no further treatment of the contacts is needed for the control synthesis.

\subsubsection{Combination of model \& learning-based locomotion}
Combinations of learning and Lyapunov-based constraints are explored in \cite{choi2020reinforcement, westenbroek2022lyapunov, meng2023hybrid}. A residual model of the robot dynamics is learned online in \cite{sun2021online}. %, and the locomotion policy adjusts to the learned model. 
% In \cite{da2019combining}, a low-dimensional manifold is learned from trajectory optimization data. 
Combinations of RL and NMPC are also proposed in \cite{melon2020reliable, yang2022fast}. % The work in \cite{melon2020reliable} used the RL output as an initial guess for the NMPC to facilitate computation and improve convergence, and the work in \cite{yang2022fast} composed RL and NMPC hierarchically, where the gait sequence output of the RL is used in the NMPC tracking controller. 
The vision of these works is to combine the strengths of model-based design and machine learning. In our work, we hope to enhance explainability of the learning-based controller with the reachability-informed design. 

\subsubsection{Physics-informed Machine Learning for Control}
The approach we undertake, which learns solutions of the HJ PDE with neural networks, falls into the category of physics-informed machine learning for control \cite{nghiem2023physics}. Solving any PDEs numerically is inherently subjected to the curse of dimensionality \cite{bellman1957}; thus, physics-informed neural network (PINN) \cite{lagaris1998artificial} is suggested as an alternative for finding approximate solutions. Its applications to solving HJ PDEs are proposed in \cite{sirignano2018dgm, darbon2020overcoming, bansal2020deepreach}, and we develop our method based on \cite{bansal2020deepreach} which is tailored for reachability.

\section{Problem Description}

\subsection{Problem Setup}
\label{Sec:2A}
\subsubsection{Hybrid dynamics model of walking robots}
% \textit{Hybrid system description of walking robot:} 
We describe the walking dynamics of the legged robot as
\vspace{-0.25em}
\begin{subequations}
\label{eq:dyn}
\begin{align}
\dot{\state} &= \dyn(\state, \ctrl), & & \prestate \notin \switchsurface \label{eq:dyn-continuous} \\
\poststate &=\resetmap\left(\prestate^{-}\right), & & \prestate^{-} \in \switchsurface, \label{eq:dyn-reset}
\end{align}
\end{subequations}
where $\state \in \R^n$ is the state and $\ctrl \in \cset$ is the control input.
$\switchsurface$ indicates the switching surface where the reset map $\resetmap$ is applied; under the reset event at time $t$, the state $\prestate^{-} :=\lim_{\tau \nearrow t}x(\tau)$ instantaneously shifts to $\poststate$. The state $\state$ consists of generalized coordinates $q$ and its time derivative; $x\!\!=\!\![q;\;\dot{q}]$. The trajectory of the robot is composed of a set of continuous trajectories driven by the continuous-mode dynamics $\dyn$, and discontinuous jumps whenever the state hits the switching surface $\switchsurface$, which captures the impact event between the swinging foot and the ground. The continuous mode dynamics \eqref{eq:dyn-continuous} can be derived from Euler-Lagrangian mechanics of the robot, constrained by the contact force at the stance leg. The reset map \eqref{eq:dyn-reset} can be modeled based on the rigid impact model and the relabeling of the coordinates for switching the swing and stance legs. For more details of the derivation of the dynamics, please refer to \cite[Sec.3.4]{westervelt2018feedback}.

\subsubsection{Stable hybrid limit cycle walking gaits} A stable walking gait of the robot is represented as a non-trivial $\period_p$-periodic solution of the system \eqref{eq:dyn}, denoted as $x^*(\cdot)$, a trajectory in time, which undergoes resets at times $k\period_p$ for positive integer $k$.
We call $\limitcycle:=\{x\;|\;\exists t\ge0, x= x^*(t)\}$ a hybrid limit cycle of the system. The limit cycle is assumed to be forward invariant and stable under some \textit{baseline controller} $\baselinectrl\!:\!\R^n\!\!\rightarrow\!\cset$, in a small neighborhood around $\limitcycle$.

\subsubsection{Parametrized walking gaits} There might exist multiple stable hybrid limit cycles of the system, each corresponding to various walking gaits of the robot. Each gait is conditioned on various gait parameters, such as average forward velocity or step length. We denote this gait parameter vector as $\gaitparam$, which will result in a parameter-conditioned hybrid limit cycle walking gait  $\limitcycle(\gaitparam)$. We will often refer to the gait parameter $\gaitparam$ itself as the ``gait'' in the manuscript for the sake of brevity. Finally, we denote $\gaitset$ as a set of gait parameters which result in a feasible stable walking gait, meaning that for all $\gaitparam \in \gaitset$, the limit cycle $\limitcycle(\gaitparam)$ exists and is stable.

\subsubsection{Unmodeled perturbations} We model perturbations not captured in \eqref{eq:dyn}, such as push or impact with other objects as a drift (or instantaneous jump) of the robot state. We assume that the perturbations we deal with are temporary. 

\vspace{-0.25em}
\subsection{Objectives}
\label{subsec:objectives}
\vspace{-0.25em}

The overall objective is twofold. First, for each gait, we seek to compute a region of state space $\RoA \subseteq \R^n$ around $\limitcycle(\gaitparam)$, from which there exists a feedback control law that asymptotically stabilizes the system to $\limitcycle(\gaitparam)$.
We call $\RoA(\gaitparam)$ the (asymptotically) stabilizable region, or \textit{region of attraction (RoA) for} $\limitcycle(\gaitparam)$. The goal is to verify in which robot state it is possible to stabilize to an individual gait, and the associated stabilizing feedback control law $\pi \!:\!\R^n\!\!\rightarrow\!\cset$.

Next, we are interested in \textit{when transitioning to a new gait is possible or necessary}, when the robot state is perturbed or if the user wants to command the robot to change its gait. That is, given a state $x$, we are interested in finding $\feasgaitset(x) \subseteq \gaitset$ such that for all gait parameters in the set, $\gaitparam \in \feasgaitset(x)$, the state is inside the RoA of the corresponding gait, $x \in \RoA(\gaitparam)$. Then, a gait transition is necessary if the current walking gait of the robot $\gaitparam$ is not in $\feasgaitset(x)$, to ensure stability of the walking. Upon transition, the robot has to select one of the gait $\gaitparam$ that is included in $\feasgaitset(x)$.

\section{Background}

\subsection{HJ reachability analysis for RoA computation}
\label{subsec:roa_hj}

In this section, we first present an overview of Hamilton-Jacobi (HJ) reachability analysis for continuous systems \cite{bansal2017hamilton}: the dynamics given by $\dot{x} = f(x, u)$ without the reset. Let $\trajstandard(\tdummy)$ denote the state at time $\tdummy$ by starting at initial state $\state$ and initial time $\tvar$, and applying input signal $\cfunc$ over $[\tvar,\tau]$. 
Given a \textit{target set} $\targetset \subset \R^n$, the \textit{Backward Reachable Tube (BRT)} of $\targetset$ is defined as the set of initial states from which there exist a control signal under which the system will eventually reach $\targetset$ within the time horizon $T$:
\vspace{-0.25em}
\begin{align*}
\footnotesize
\brt(\targetset; \horizon)\!:=\!\{\state \; | \; & \exists \ctrl\!:\![-\horizon, 0]\rightarrow\cset, \\
& \exists \tdummy\!\in\![-\horizon, 0], \traj_{\state,-T}^{\ctrl}(\tdummy)\!\in\!\targetset\}. \normalsize
\vspace{-0.25em}
\end{align*}

For our problem, we define the target set as a small neighborhood of the gait $\limitcycle(\gaitparam)$, denoted as $\targetset(\gaitparam)$, such that the baseline controller $\pi_0$ can stabilize to the limit cycle from anywhere inside $\targetset$. Such a neighborhood region exists for any asymptotically stable attractor \cite{clf_zubov}. If we set $\limitcycle(\gaitparam)$ directly as the target set, only the states that can achieve \textit{finite-time} convergence to the gait can be verified, excluding states that are not finite-time stabilizable but still asymptotically stabilizable to the gait. By the definition of BRTs, any state $x$ in $\brt(\targetset(\gaitparam); T)$ is reachable to $\targetset(\gaitparam)$ in finite-time, and once it reaches the target set, it can be stabilized to $\limitcycle(\gaitparam)$.
Therefore, for every state that can be verified as an element of a finite-time BRT of $\targetset(\gaitparam)$, we can conclude that it is an element of $\RoA(\gaitparam)$, the RoA of the gait. In fact, if $\horizon \rightarrow \infty$, the BRT recovers the full RoA of the given gait, i.e. $\lim_{\horizon \rightarrow \infty} \brt(\targetset(\gaitparam); T) = \RoA(\gaitparam)$. Thus, the BRTs computed for long enough time horizon can be considered as a maximal estimation of the RoAs of the gaits. 

In HJ reachability, the computation of BRT is considered an optimal control problem, which can be solved with dynamic programming. First, a signed distance function to $\targetset$, $\targetfunc(\state)$, is defined whose zero-sublevel set is $\targetset$, i.e. $\targetset = \{\state : \targetfunc(\state) \leq 0\}$. Here, the dependency on $\gaitparam$ is dropped for simplicity. Next, we define the minimum signed distance to $\targetset$ over time along the trajectory as
\vspace{-0.25em}
\begin{equation}
    \label{eq:costfunctional}
    \costfunctional(\tvar, \state,\cfunc) = \min_{\tdummy \in [-\tvar, 0]} \targetfunc(\traj_{\state,-\tvar}^{\ctrl}(\tdummy)).
\vspace{-0.25em}
\end{equation} 
When $\targetfunc(\tdummy)\le0$, the system is inside $\targetset$ at time $\tdummy$, thus, reaches the target set. Thus, for the goal of computing the BRT, we compute the optimal control that minimizes this distance (so that it achieves a non-positive value of $\costfunctional$). As such, we define the value function as
\vspace{-0.25em}
\begin{equation}
    \label{eq:valuefunc}
    \vfunc(\tvar, \state) = \inf_{\cfunc} \Big\{\costfunctional\Big(\tvar,\state,\cfunc\Big)\Big\}.
\vspace{-0.25em}
\end{equation}
The value function in \eqref{eq:valuefunc} can be computed using dynamic programming, which results in the following Hamilton-Jacobi partial differential equation (HJ PDE) \cite{fisac2015reach}:
\vspace{-0.25em}
\begin{equation}
\begin{aligned}
    \label{eq:HJIVI}
    \min\big\{\!-\!D_\tvar \vfunc(\tvar,\state)+ H(\tvar,\state), \; \targetfunc(\state)-\vfunc(\tvar,\state)\big\} = 0,
    \end{aligned}
\vspace{-0.25em}
\end{equation}
with the initial value function $\vfunc(0, \state) = \targetfunc(\state)$, where
\vspace{-0.25em}
\begin{equation}
    \label{eq:ham}
    \begin{aligned}
    H(\tvar,\state) := \min_\ctrl & \nabla \vfunc(\tvar,\state) \cdot \dyn(\state,\ctrl).
        \end{aligned}
\vspace{-0.25em}
\end{equation}
$D_\tvar$ and $\nabla$ represent the time and spatial gradients of the value function. 
$H$ is the Hamiltonian that encodes the role of dynamics and the optimal control input. 
% Intuitively, the term $\targetfunc(\state)-\vfunc(\tvar,\state)$ in \eqref{eq:HJIVI} enforces the value function to ``memorize'' the best record (closest instance to the target set) of the optimal trajectories. 
Once $\vfunc(\tvar,\state)$ is obtained by solving the HJ PDE, the BRT is given as the zero sub-level set of the value function
$\brt(\targetset(\gaitparam); t) = \{\state \; | \; \vfunc(\tvar,\state) \leq 0 \}$. The corresponding optimal control for reaching the target set $\targetset$ is derived as
\begin{equation}
    \label{eq:opt_ctrl}
    \begin{aligned}
    \pi^*(\tvar,\state) = \arg\min_\ctrl \nabla \vfunc(\tvar,\state) \cdot \dyn(\state,\ctrl).
        \end{aligned}
\end{equation}

\subsection{HJ reachability for walking robots}
We next summarize the extension of the reachability framework in \cite{choi2022}, to account for discontinuous state resets. The value function in the presence of state resets, \eqref{eq:dyn-reset}, can be obtained by solving a constrained version of the HJ PDE:

\small
\vspace{-1.25em}
\begin{subequations}
\label{eq:HJIVI_constrained}
\begin{align}    
    \min\!\Big\{\!\!-\!D_\tvar \vfunc(\tvar,\state)\!+\!H(\tvar,\state), \;\targetfunc(\state)\!-\!\vfunc(\tvar,\state)\Big\}\!=\!0, & & \prestate \notin \switchsurface, \label{eq:HJ PDE-continuous} \\
    \vfunc(\tvar,\state) = \vfunc(\tvar, \resetmap\left(\prestate\right)), & & \prestate \in \switchsurface, \label{eq:value-remapping}
    \end{align}
\end{subequations}
\normalsize
\vspace{-1.5em}

\noindent with the initial value function 
\vspace{-0.25em}
\begin{equation}
\label{eq:HJIVI_constrained_IC}
    \vfunc(0, \state)\!=\!\targetfunc(\state) \; \text{if}\; \prestate\!\notin\!\switchsurface, \quad \vfunc(0, \state)\!=\!\targetfunc(\resetmap\left(\prestate\right)) \; \text{if}\; \prestate\!\in\!\switchsurface.
\vspace{-0.25em}
\end{equation}
In words, the value function can be obtained by solving the usual HJ PDE for the states that are not on the switching surface, and for the states that are on the switching surface, the value is given by that of the corresponding post-reset state.
This is because if the state is on the switching surface, it will instantaneously change to the post-reset state. Since \eqref{eq:HJIVI_constrained} reasons about the state resets, the obtained value function and the associated optimal controller \eqref{eq:opt_ctrl} implicitly exploit the reset map to reach the target set as quickly as possible.

The HJ PDEs in \eqref{eq:HJIVI} and \eqref{eq:HJIVI_constrained} can be effectively solved using numerical algorithms like level set methods described in \cite{fisac2015reach, mitchell2004toolbox}. The only additional step in solving \eqref{eq:HJIVI_constrained}, compared to \eqref{eq:HJIVI} is to enforce \eqref{eq:value-remapping} at every timestep, which does not add computational complexity to the original algorithm. Please refer to \cite{choi2022} for more details of the numerical algorithm.

\subsection{Limitations of numerical methods for HJ reachability}
\label{subsec:limitation_numerical}
Applying the numerical algorithms for solving the HJ PDE to legged robot dynamics encounters several key obstacles. First, since the algorithm is in essence a brute force dynamic programming, it is practically infeasible to be applied to realistic legged robots due to the curse of dimensionality \cite{bellman1957}. The computational load and the memory requirements grow exponentially with respect to the state dimension. 

Next,  the numerical stability of the PDE solutions is tightly coupled with the maximal norm of the Hamiltonian \eqref{eq:ham} \cite{Mitchell02}. This necessitates a denser computational grid for robots with stiffer dynamics, characterized by larger magnitudes of their vector fields. Legged robots exhibit stiffer behaviors than simpler systems like mobile robots or near-hover drones \cite{bansal2017hamilton}, previously addressed by HJ reachability. Consequently, balancing computational time against numerical accuracy often results in a value function whose gradient, critical for determining the optimal control, suffers poor accuracy.

Finally, the computation of BRTs for legged robots presents another efficiency challenge when considering the gait parameter $\gaitparam$.
Solving for BRTs individually across each gait parameter leads to considerable computational redundancy and memory waste. This is because BRTs for neighboring gaits tend to exhibit similar shapes, as we will see later in the paper. In this regard, a parametric approach to representing the gait BRTs can address these inefficiencies effectively. By computing the BRTs in a unified process that accounts for all possible gait parameters, only the shape parameters of the BRT need to be stored and can significantly reduce the computation and memory requirements.

\subsection{Deep learning-based reachability for continuous systems}

In light of the limitations of the brute force numerical methods in solving the HJ PDE, a deep learning-based approximate solution for HJ reachability, named DeepReach, was proposed in \cite{bansal2020deepreach} for continuous systems. DeepReach utilizes sinusoidal activation functions \cite{sitzmann2020implicit} to represent the value function and employs a loss function that learns the HJ PDE solution in self-supervised fashion. 

During the training, DeepReach samples a batch of time and state samples from the target domain. The loss function for a given sample $(\statetimesample)$ where $i$ is the index of the sample, is given as $h(t_i,x_i) = \lambda_1 h_1 + \lambda_2 h_2$ where 
\vspace{-0.25em}
\begin{equation}
\label{eq:loss_terms}
\begin{aligned}
h_1 = &\big| \min\{-\!D_tV_{\theta}(\statetimesample)+H(\statetimesample), \\
& \quad \quad\;\;\; l(x_i)-V_{\theta}(t_i,x_i) \}\big|,\\
h_2 = & |l(x_i)-V_{\theta}(0, x_i)|.
\end{aligned}
\vspace{-0.25em}
\end{equation}
$h_1$ evaluates the left hand side of \eqref{eq:HJIVI} at the training sample, and $h_2$ evaluates the initial condition of the PDE.

Since $h_1$ depends on gradients of the value function, the neural network should not only approximate the value function well but also its gradients. The widely popular ReLU-based neural networks struggle to accurately represent their gradients, which can lead to a poor approximation of the value function. Thus, DeepReach employs a sinusoidal activation function \cite{sitzmann2020implicit}, which is known to produce accurate gradients due to its inherent differentiability. After the training, the BRT can be represented by the zero-sublevel set of the learned value function.

\section{Our Method}

\subsection{Extension of DeepReach for learning gait BRTs}

Our method mainly extends the DeepReach framework to address the parameter-conditioned varying target gaits and the hybrid dynamics of the legged robots. In our framework, we incorporate four key features not present in the original DeepReach work: 1) parameterization of the value function with respect to the gait parameter $\gaitparam$, 2) an additional loss term that captures the condition \eqref{eq:value-remapping} resulting from the state reset, and 3) sequence-to-sequence (Seq2seq) training scheme, to mitigate the ``forgetting'' effect in long-horizon training. We provide the details of each extension below.

\subsubsection{Extension of DeepReach to parameterized BRTs of hybrid limit cycle}

The neural network value function will be denoted as $V_{\theta}(\tvar,\state)$, where $\theta$ indicates the weights of the sinusoidal network. As proposed in \cite{borquez2023parameter}, we can augment the input of the network to treat the gait parameter as a virtual state. Consequently, our value function is expressed as $V_{\theta}(\tvar,\state;\gaitparam)$. This enables us to derive RoAs for different gaits using a single learned value function model and through a single training procedure.

% loss function
\subsubsection{Loss function} In each training iteration, we sample a batch of time, state, and gait parameter samples from the target domain of each input entity.
The loss function for a given sample $(\statetimegaitsample)$, consists of four loss terms as below:
\vspace{-0.25em}
\begin{equation}
    h(t_i,x_i;\gaitparam_i) = \lambda_1 h_1 + \lambda_2 h_2 + \lambda_3 h_3 + \lambda_4 h_4, 
\end{equation}
\vspace{-0.25em}
where $h_1$ and $h_2$ are given in \eqref{eq:loss_terms}, and
\vspace{-0.25em}
\begin{align}
% h_1 = &\big| \min\{-\!D_tV_{\theta}(\statetimegaitsample)+H(\statetimegaitsample), \label{eq:loss_terms}\\
% & \quad \quad\;\;\; l(x_i;\gaitparam_i)-V_{\theta}(t_i,x_i;\gaitparam_i) \}\big|, \;\; \text{for} \; x_i\notin S,  \nonumber\\
% h_2 = & |l(x_i;\gaitparam_i)-V_{\theta}(0, x_i ;\gaitparam_i)|,   \nonumber\\
h_3 = & |V_{\theta}(\statetimegaitsample)-V_{\theta}(t_i,\Delta(x_i);\gaitparam_i)|\; (\text{for} \;x_i \in S), \label{eq:loss_terms2}\\
h_4 = & \max\{V(t_i,x_i,\gaitparam_i)-l(x_i;\gaitparam_i),0\}\nonumber\\
&+ \max\{V(t_i,x_i,\gaitparam_i)\!-\!V(t_j,x_i,\gaitparam_i),0\} \;\; \text{where} \; t_j\!<\! t_i. \nonumber
\vspace{-0.25em}
\end{align}
Here, $\lambda_i$ for $i=1,2,3,4$ are weights of each loss term. We introduce two new loss terms, $h_3$ and $h_4$. The loss term $h_3$ ensures that the states before and after the reset share the same value function values, to satisfy the condition \eqref{eq:value-remapping}. This is the essential loss term that captures the effect of the impact event of the walking behavior to the gait stabilization. 
From the definition of the value function in \eqref{eq:valuefunc}, it can only monotonically decrease in time due to $\min_{\tdummy \in [-\tvar, 0]}$ in \eqref{eq:costfunctional}. Thus, we impose this condition by adding the loss term $h_4$, which penalizes $V_\theta$ if the monotonically decreasing condition is not met. Including this term incentivizes the neural network to learn a more accurate value function.
% learn the target function

\subsubsection{Seq2seq Training}
In DeepReach, during the training, the sample time domain $[0, T]$ is scheduled in a curriculum learning fashion, by gradually increasing the maximum time $T$. It is important that the value function is shaped from the initial condition constraint in the beginning, and it is carved out as the training proceeds through the PDE loss $h_1$ and other loss terms. Thus, the initial condition serves as an ``anchor'' for the value function.
However, the anchoring effect of the initial condition gets less effective for $t$ that is further away from 0. Thus, an apparent issue with training a single model for the entire time domain $[0, T]$ is that the value function starts to forget information of $l$ when the time horizon is longer. % The common phenomenon of this ``forgetting'' is that the shape of the BRTs get more blurry for larger $t$, losing details of the actual shape. 
This issue is more severe when the dynamics are stiff and involve state jumps, as in our problem. % A similar issue is investigated in the deep RL literature in the absence of corrective feedback \cite{kumar2020discor}.

To mitigate this issue, we employ a Seq2seq training scheme from \cite{krishnapriyan2021characterizing}. Basically, Seq2seq splits the time domain into multiple subdomains, and trains separate neural network models for each subdomain. This divide-and-conquer approach is effective in mitigating the forgetting phenomenon; for each subdomain, we can introduce the initial condition loss again, which will anchor the value function not only at $t=0$ but also at the intervals of the subdomains. % The entire time horizon is divided into multiple sequences, and 
Each subsequent sequence benefits from the model trained on the preceding sequence as its supervision signal. 

\subsection{One-step predictive stabilizing controller}
\label{sec:4B}
Once we train the value function \( V_{\theta}(t,x;\beta) \), we can easily derive the gradient-based optimal controller in \eqref{eq:opt_ctrl}, to stabilize the states within the RoA to the gait. However, in our evaluation, applying \eqref{eq:opt_ctrl} was not successful mainly due to the learning errors of the value function and its gradients. The effect of the error is severe for the closed-loop performance of the gradient-based controller. This is because when the trajectory evolves, the accumulation of the error effect is tightly coupled with the stability of the closed-loop dynamics. It becomes a more severe issue for legged robots that involve discrete contact dynamics \eqref{eq:dyn-reset}.

In this work, we design our controller as an one-step predictive (OSP) control problem, which does not directly rely on the value function gradient. This formulation is in fact the discrete-time approximation of the optimal control law in \eqref{eq:opt_ctrl}, and if the value function is accurate, they should produce the same result with small enough timestep. However, with the neural network value function, in our experiments, we observed that the new formulation achieves a much higher success rate of stabilization than \eqref{eq:opt_ctrl}. 
At each time step $i$, the OSP controller solves an optimization problem
\begin{equation}
\begin{aligned}
    \label{eq:osp_problem}
    \min_{u_i \in \cset} & V_{\theta}(t_i,\hat{x}_{i+1};\beta) \\
     % &\quad s.t. \quad V_{\theta}(T,x_{i+1};\beta_{i+1})\le 0\\
    \text{where} \; & t_i \; \text{is such that} \; V_{\theta}(t_i,x_{i};\beta) = 0, \\
    &\hat{x}_{i+1}=\mathrm{f}(x_i, u_i).
    \end{aligned}
\end{equation}

The time $t_i$ is determined as a (minimal) \textit{Time-to-Reach (TTR)} at the current state $x_i$ \cite{Mitchell02}, clipped by the time horizon $T$. Since the value function is non-increasing with respect to time, such TTR value is uniquely determined. We can find the value by doing a binary search for when $V_{\theta}(t_i,x_{i};\beta)$ becomes zero. By taking TTR as the time index for the value function evaluation, the controller is trying to ``slide along with'' the zero-level set of the value function---the BRT---over time, until the BRT shrinks to the target set. 
% This is what the ground-truth optimal trajectory would do, if it is controlled under the original optimal control policy \eqref{eq:opt_ctrl}.
The discrete-time dynamics $\mathrm{f}(x, u)$ in \eqref{eq:osp_problem} is determined based on the continuous system dynamics described in \eqref{eq:dyn}. % It can be readily obtained using numerical integration such as Euler or Runge-Kutta methods. 
Finally, the objective determines a control input that leads the state at the next time step to the one where the value function is maximally decreased. By iteratively updating the TTR and descending the value function, the controller can ultimately converge to the target set.

Given that the optimization problem involves a non-convex cost function and the constraint represented by a neural network, one can adopt a discretization or sampling-based search (evaluating over discrete samples in the input set $\cset$) to perform the optimization. For our one-step problem, this can be done promptly in real-time as it only requires a single-instance batch neural network inference for all the samples. However, the complexity can increase if \eqref{eq:osp_problem} is extended to a multi-step predictive control problem.

\begin{figure}[t]\centering
\includegraphics[width=\columnwidth]{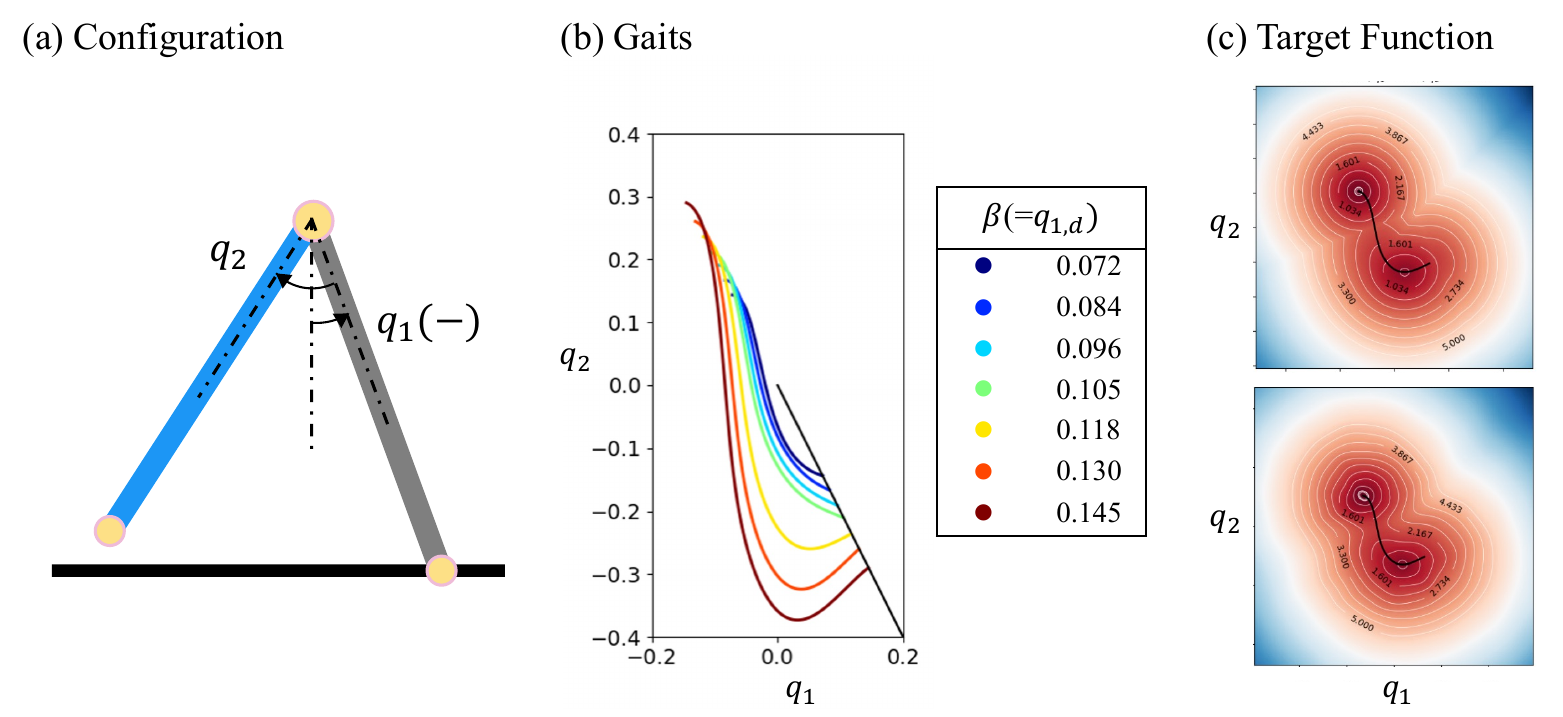}
\vspace{-1.5em}
\caption{(a) Configuration of the two-link walker (grey: stance leg, blue: swing leg). (b) Hybrid limit cycle gaits in $q_1$-$q_2$ space with various walking step lengths. Black line indicates the switching surface. (c) $q_1$-$q_2$ slice of the numerical (top) and learned (bottom) target function $l(x;\gaitparam)$ for $\gaitparam=0.13$.}
\vspace{-0.75em}
\label{fig:config}
\end{figure}

\begin{figure}[t]
\centering
\includegraphics[width=\columnwidth]{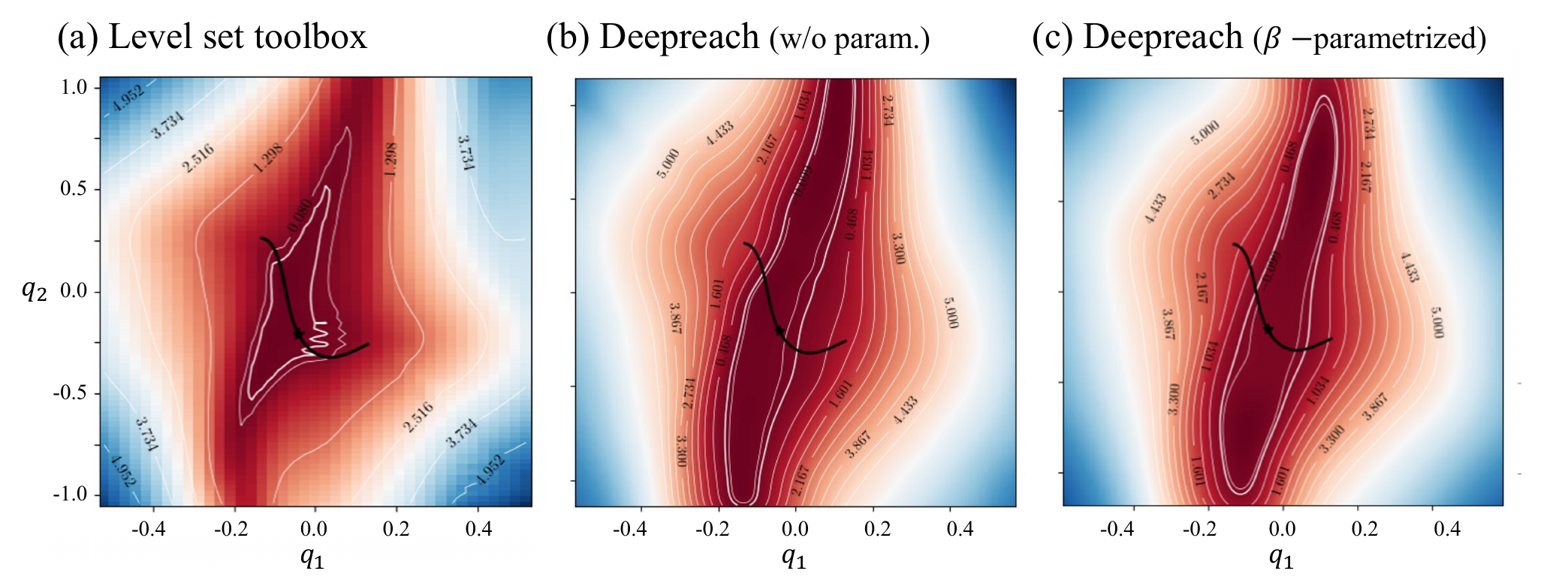}
\vspace{-1.5em}
\caption{Comparison between the value functions $V(T,x;\beta)$ for the gait with $\beta\!=\!0.13$, obtained by (a) the numerical method in \cite{mitchell2004toolbox, choi2022}, and (b) (c) our method, where in (b), $V_\theta$ is not parametrized and $\beta$ is fixed as 0.13. The values are visualized in color contour map in $q_1-q_2$ slices along the gait. The zero-level sets (thick white line) represent the estimated BRTs.}
% \vspace{-1em}
\label{fig:value}
\end{figure}

% \begin{remark} \eqref{eq:osp_problem} has resemblance to the optimal policy in fitted value iteration \cite{sutton2018reinforcement}, although \eqref{eq:osp_problem} is specific for finite-time reachability problem. %, whereas the classic fitted value iteration is commonly applied to infinite-horizon accumulative reward problems.
% \end{remark}

\subsection{Gait switching controller}
\label{subsec:switching_ctrl}
\setlength{\textfloatsep}{0pt}% Remove \textfloatsep
\begin{algorithm}[t]
\SetAlgoLined
\SetKwInOut{Input}{Input}
\SetKwInOut{Output}{Output}
\caption{Gait switching strategy}
\label{algorithm1}
\Input{Current state $x_i$, Desired gait parameter $\gaitparam^*$}
\Output{Selected gait parameter $\beta_{i}$}
\If{$V_{\theta}(T,x_i;\beta^*)\le 0$}{$\textbf{return\ } \beta^* $}
$\mathrm{min\_value} \gets +\infty$,~~$\beta_{i} \gets \textbf{None}$

\For{$\beta$ in $\gaitset$}{
\If{$\mathrm{min\_value}>V_{\theta}(T,x_i;\beta) $}{
$\mathrm{min\_value}\gets V_{\theta}(T,x_i;\beta)$,~~$\beta_{i} \gets \beta$
}
}
\If {$\mathrm{min\_value}>0$}{Warning: $\feasgaitset(x_i)$ is empty}
$\textbf{return } \beta_{i}$
\end{algorithm}

Based on the OSP controller in \eqref{eq:osp_problem}, we are able to effectively stabilize the robot state to a single gait $\limitcycle(\beta)$, when the state $x_i$ is inside $\brt(\targetset(\beta); T)$, where $T$ is the maximum time horizon the value function is trained for. However, if $x_i \notin \brt(\targetset(\beta); T)$, although in practice we can still deploy \eqref{eq:osp_problem} by setting the TTR $t_i$ as $T$, since the state is outside of the RoA estimate, there is no guarantee that it will eventually stabilize to the gait. Recalling the second objective in Section \ref{subsec:objectives}, we always want to commit to a gait to which stabilization is feasible. The main advantage of having access to the parametrized value function is being able to switch the target $\beta$ actively online to achieve this. 

We assume that a desired gait parameter $\beta^*$ is specified by a user command or by a pre-specified sequence of desired gaits. Whenever before \eqref{eq:osp_problem} is executed with $\beta=\beta^*$, we can check whether $x_i \in \brt(\targetset(\beta^*); T)$, and if this condition is not satisfied, change $\beta$ to a member of $\feasgaitset(x_i)$. In the case where the state $x_i$ is perturbed by unmodeled disturbance, the feasibility condition will be checked for all possible gaits and the gait will be switched to a new feasible gait. For perturbations whose magnitude is significant so that $\feasgaitset(x_i)$ is empty, the user will be aware that the walking stability is not ensured anymore. This gait switching strategy is summarized in Algorithm \ref{algorithm1}.

\begin{figure*}[h]
    \centering
\includegraphics[width=0.9\textwidth]{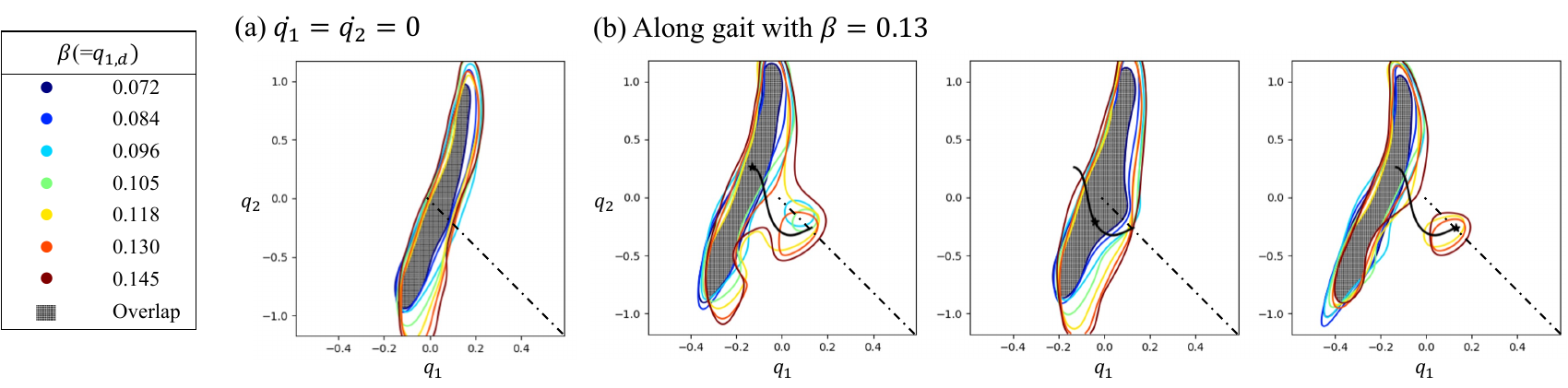}
\vspace{-0.5em}
    \caption{$q_1$-$q_2$ slices of gait BRTs (a) when angular velocities are 0, (b) values that lie on the gait with $\beta=0.13$ (* indicates where the slice is taken). Each BRT captures states from which the robot can be stabilized to the corresponding gait, and in the overlap region, pursuing any gait is feasible.}
    \label{fig:brt_overlap}
\vspace{-1.25em}    
\end{figure*}

\section{Case Study: Two-link Walker}

\begin{figure}\centering
\includegraphics[width=\columnwidth]{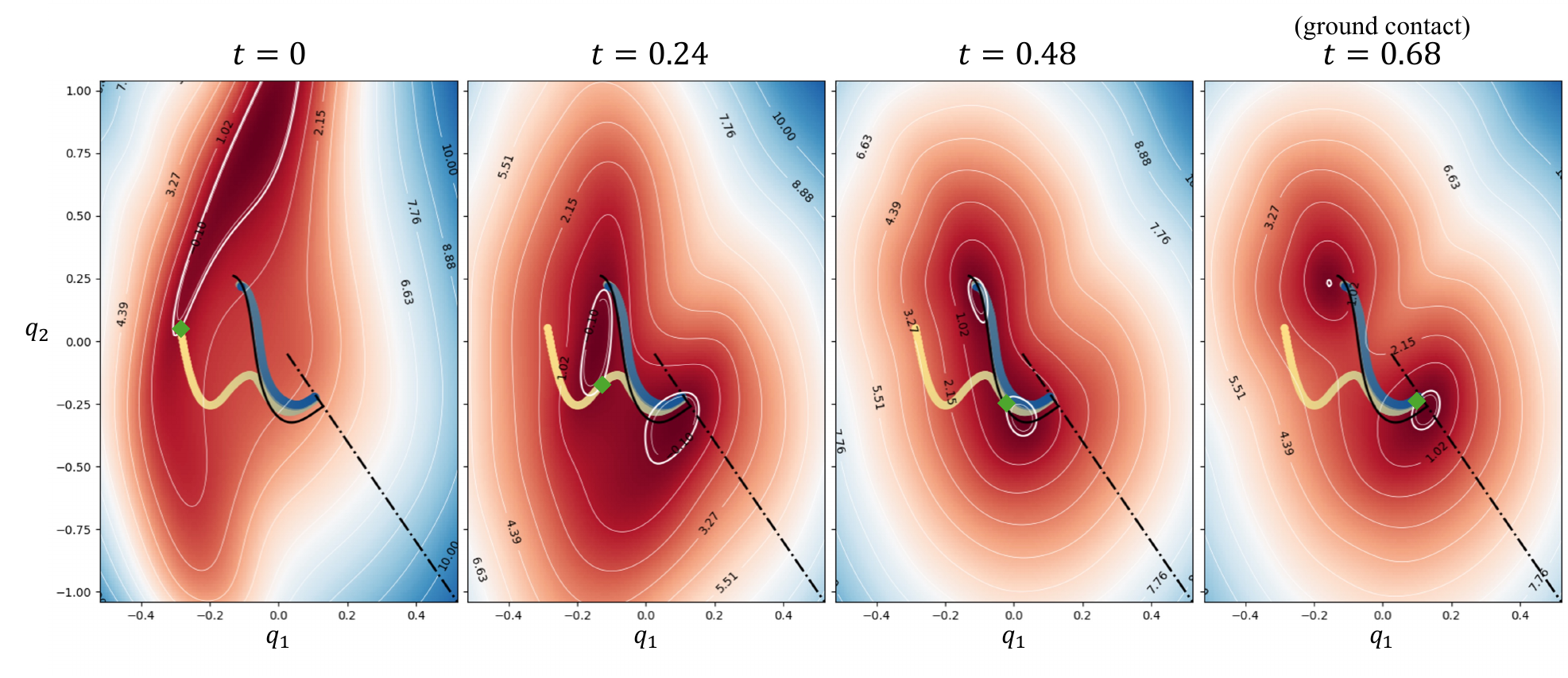}
\vspace{-1.25em}
\caption{Snapshots of the phase portrait of the trajectory, initialized at a perturbed state, stabilizing to the gait with $\beta=0.13$, under the OSP controller \eqref{eq:osp_problem}. The trajectory evolves from yellow to blue while taking two walking steps, and we show the first walking step portion. Green dots represent the state where the snapshot is taken. The color contour map visualizes $V_\theta(t_i, \cdot;\beta)$ in \eqref{eq:osp_problem}. [\href{https://youtu.be/P7Vnr8jwSPc}{Video} (\url{https://youtu.be/P7Vnr8jwSPc})]}
\label{fig:snapshot}
\vspace{-0.5em}    
\end{figure}
% analyze the BRT result

We consider a compass-gait walker, which consists of two links with an actuated joint between them. We consider a pinned model of the robot, with the configuration variable $q \coloneqq [q_1, q_2]^T$ as illustrated in Fig. \ref{fig:config} (a), and we define the state as $\state \coloneqq [q, \dot{q}]^T$. The switching surface is defined as where the swing foot hits the ground with a negative velocity and the stance leg angle crosses a predefined threshold $\bar{q}_1$, $\switchsurface\!:=\!\lbrace \state ~|~ q_1\!\leq\!\bar{q}_1, 2q_1\!+\!q_2\!=\!0, 2\dot{q}_1\!+\!\dot{q}_2\!<\!0 \rbrace.$

The stable gaits of the robot are represented as swing leg angle $q_{2,r}$ being a polynomial function of $q_1$. These polynomial gaits are obtained from trajectory optimization \cite{westervelt2018feedback} for desired stance leg angle at the event of impact, $q_{1,d}$, ranging from 0.072 to 0.145. We set $\beta=q_{1,d}$ which decides the walking step length of the gait. The closed-loop dynamics under an input-output (IO) linearization controller \cite{sreenath2011compliant} is considered in the optimization, thus, the obtained gaits are stable limit cycle under the IO linearization controller. The IO linearization controller also provides the baseline stabilzing controller $\pi_0$ discussed in Section \ref{Sec:2A}. The obtained parametrized gaits are visualized in Fig. \ref{fig:config} (b).

\subsubsection{Training details}

The target function $l(x;\gaitparam)$ is constructed by evaluating the distance between the state $x$ and the gait $\limitcycle(\gaitparam)$. Evaluating the distance numerically for all samples in each training iteration significantly slows down the training. Instead, we use a neural network to represent $l(x;\gaitparam)$. We generate 100,000 samples with 21 values of $\gaitparam$ from $\mathcal{X}$ and $\gaitset$, and learn the target function with supervised learning. The learned $l(x;\gaitparam)$ is shown in Fig. \ref{fig:config} (c).

In the value function training, we employ a 3-layer neural network with 512 hidden nodes in each layer to represent the learned value function and we utilize the sinusoidal function as the activation function. Additionally, we set the time span of the BRTs to $T=0.5$. We break up the time span to four sequences for Seq2seq training. In each sequence, we uniformly sample 130,000 samples of $(\statetimegaitsample)$. It takes approximately 8 hours to complete one sequence of training on the RTXA5000 GPU, and total time cost for the entire training of the parametrized BRTs is 32 hours. This is notably shorter than the direct numerical method in \cite{choi2022}, as a computation for a \textit{single} gait BRT takes 12 hours.

\subsubsection{Learned Value function} 

The learned value function is visualized in Fig. \ref{fig:value} for $\beta=0.13$, where we compare our solution to the numerical solution obtained in \cite{choi2022}. Note that the numerical solution is not necessarily the ``ground-truth'', as discussed in Sec. \ref{subsec:limitation_numerical}, due to numerical errors. The value functions of our method are calibrated after training, based on the approach in \cite{lin2023generating}, which provides an empirical 95\% success rate of stabilization from the calibrated BRT. Overall, our learned value function generates larger estimation of the BRT compared to the numerical solution. Meanwhile, our parametrization does not sacrifice much accuracy in BRT estimation, as depicted in Fig. \ref{fig:value} (b) and (c).

\subsubsection{Regions of Attraction}
The trained BRTs, representing the RoAs of the gaits, given as the zero-level sets of the learned value function $V_\theta(T,x;\beta)$, are visualized in Fig. \ref{fig:brt_overlap}. For any states encapsulated in the BRT of $\gaitparam$, theoretically, we can ensure their convergence to the target gait of parameter $\gaitparam$. From the overlapping region of all BRTs, any gait parameter can be selected as the target gait. In contrast, at a state that does not belong to any RoAs, it might not be feasible to stabilize the robot, no matter how good the control policy is.

\begin{figure}\centering
\includegraphics[width=\columnwidth]{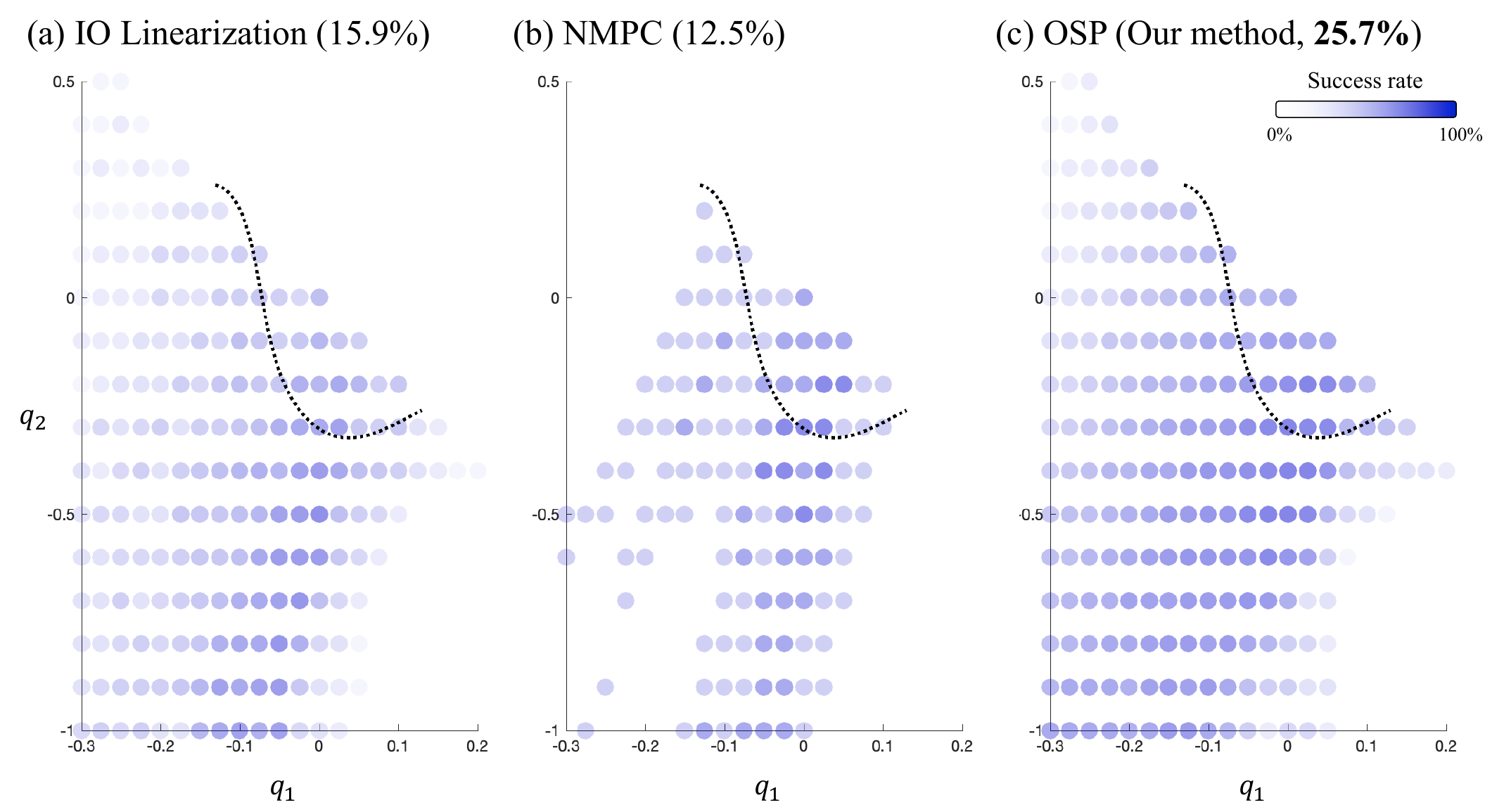}
\vspace{-1em}
\caption{Success rate of stabilization evaluated over a grid of 6,600 initial states. At each $(q_1, q_2) \in [-0.3, 0.3]\times[-1.0,1.0]$ value, we evaluate 25 combinations of $(\dot{q}_1, \dot{q}_2) \in \times[-1.0, 1.0]\times[-2.5, 2.5]$, and visualize the rate of the trajectories successfully converging to the gait ($\beta=0.13$).}
\label{fig:monte-carlo}
% \vspace{-0.5em}    
\end{figure}

\subsubsection{Stabilization performance and comparison to other controllers} The mechanism of the OSP controller in \eqref{eq:osp_problem} is visualized in Fig. \ref{fig:snapshot}, along a trajectory that is stabilized from a perturbed initial state to the gait. It shows that over time, the BRT evaluated at TTR shrinks to the target set and guides the trajectory to successfully converge to the gait. Next, we do a quantitative analysis of the performance of our stabilization controller. We evaluate the success rate of the stabilization within two walking steps among 6,600 trajectories initialized within the state space grid. The states whose swing foot lies below the ground are filtered out, as they represent physically unrealistic configurations. The learned gait BRT encapsulates $13.74\%$ of the tested initial states. We compare the success rates of our controller, the IO linearization controller, and a simple receding-horizon NMPC controller that minimizes the tracking error to the gait. The NMPC prediction horizon is set to 0.35, and increasing the horizon decreased the success rate, due to more occurrence of infeasible solutions. The success rates and the hit map of the successful initial states are reported in Fig. \ref{fig:monte-carlo}. The success rate of our controller ($25.7\%$) surpasses the success rates of the others by $10\%$. 

\subsubsection{Gait switching for enhanced stabilization}
\begin{figure*}[h]
    \centering
\includegraphics[width=\textwidth]{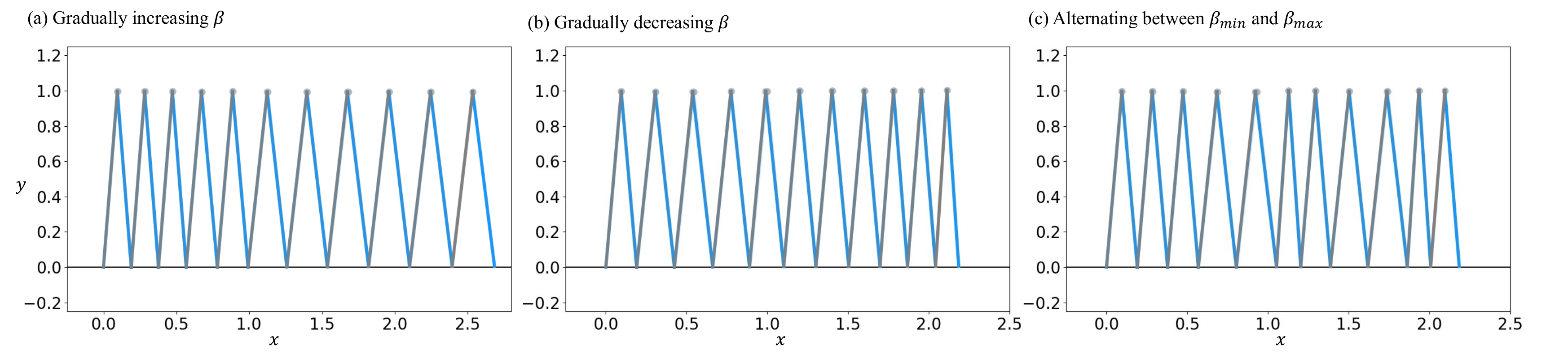}
\vspace{-1.75em}
    \caption{Two-link walker trajectory snapshots under the gait switching strategy in Algorithm \ref{algorithm1} and the OSP controller, when the sequence of gait is commanded. (a) gradually increasing $\beta$, (b) gradually decreasing $\beta$, and (c) switching $\beta$ between its min and max values for every two steps. [\href{https://youtu.be/P7Vnr8jwSPc}{Video}]}
    \label{fig:switching_snapshot}
\vspace{-1.5em}
\end{figure*}

When the user commands to change the gait or if a perturbation occurs to our robot, it becomes necessary for the robot to transition and stabilize to a new gait based on the gait switching strategy in Algorithm \ref{algorithm1}. We first demonstrate scenarios where the user commands a sequence of desired gaits. We evaluate our controller under the three different commands of desired gait sequences: (a) gradually increasing $\beta$, (b) gradually decreasing $\beta$, and (c) dramatically switching $\beta$ between its minimum and maximum values for every two steps. The results are displayed in Fig. \ref{fig:switching_snapshot}. The proposed algorithm enables the robot to switch between different gaits while trading off stability against the commanded gait. This can be particularly seen in the third case where the robot doesn't immediately switch to the minimum gait as that might result in loss of stability---instead, the robot switches to intermediate gaits determined by the algorithm.

\begin{figure}\centering
\includegraphics[width=\columnwidth]{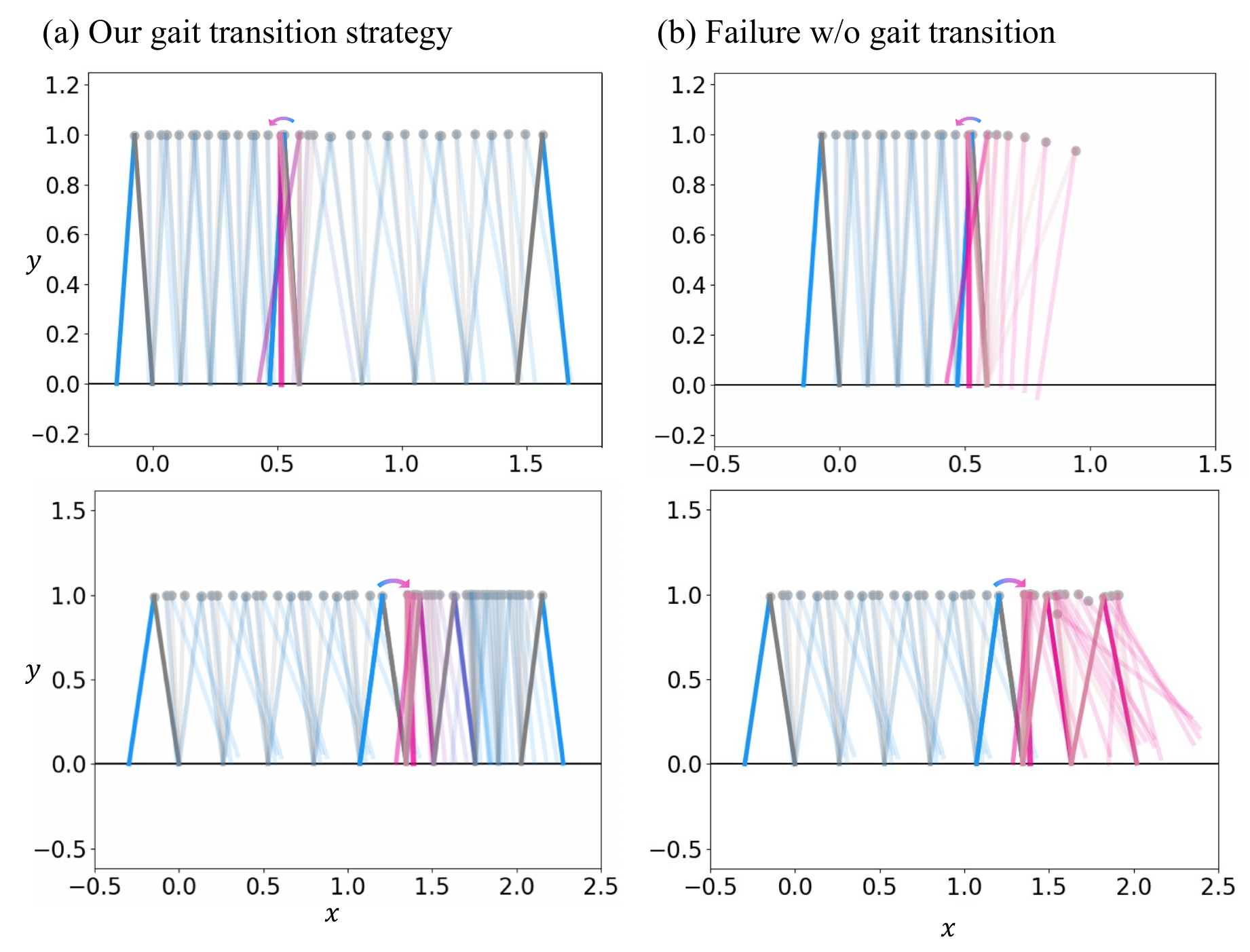}
\vspace{-1.75em}
\caption{Strong perturbations (red) causing the robot state to exit the BRT of the initial gait $\beta_1$ (top: $\beta_1\!=\!0.096$, bottom: $\beta_1\!=\!0.1485$) (a) Algorithm \ref{algorithm1} effectively switch the gait and prevents the robot from falling, whereas (b) maintaining the initial gait leads to failure. [\href{https://youtu.be/P7Vnr8jwSPc}{Video}]}
\label{fig:perturbation}
% \vspace{-0.em}
\end{figure}

We also introduce a strong perturbation to the robot which is stably tracking an initial gait ($\beta_1$), mandating it to transition to a new gait ($\beta_2$) since the perturbed state is not included in the $\brt(\beta_1)$. We utilize our gait switching controller to identify a new feasible gait and stabilize the robot to it. The results are shown in Fig.\ref{fig:perturbation}.

\section{Conclusion}

In this work, we utilized deep learning-based reachability analysis to create a library of Regions of Attraction (RoAs) for various gaits of legged robots with hybrid dynamics. The analysis with the estimated RoAs provides a transparent logic behind our gait-stabilizing controller and the gait switching strategy.

In future research, several intriguing directions are worth exploring. First, although our use of neural networks to approximate solutions to the HJ PDE has shown promise, our learned value function can be still inaccurate due to accumulated errors in the learning process. Therefore, the estimated RoAs cannot precisely guarantee safety for the robots. Investigating recent advancements that provide probabilistic guarantees on learned solutions \cite{lin2023generating} could mitigate this limitation. Additionally, we aim to explore scenarios with persistent disturbances, such as payloads, and design control policies that are robust against bounded disturbances. This could be achieved by employing a differential game-based robust reachability formulation \cite{fisac2015reach}. Finally, applying our approach to higher-dimensional, real-world walking robots will be an exciting direction.

\bibliographystyle{IEEEtran}
\bibliography{references}
\balance

\end{document}

%% file: notations.tex
\newcommand{\tvar}{t}
\newcommand{\tdummy}{\tau}
\newcommand{\R}{\mathbb{R}}
\newcommand{\ctrl}{u}

\newcommand{\cfunc}{u(\cdot)}

\newcommand{\cset}{\mathcal{U}}

\newcommand{\state}{x}
\newcommand{\traj}{\xi} 

\newcommand{\dyn}{f} 
\newcommand{\targetfunc}{l}
\newcommand{\targetset}{L}

\newcommand{\costfunctional}{J}

\newcommand{\vfunc}{V}

\newcommand{\trajstandard}{\traj_{\state,\tvar}^{\ctrl}}

\newcommand{\prestate}{\state}
\newcommand{\poststate}{\state^{+}}
\newcommand{\switchsurface}{S}
\newcommand{\resetmap}{\Delta}
\newcommand{\limitcycle}{O}

\newcommand{\period}{T}
\newcommand{\RoA}{\Omega}

\newcommand{\horizon}{T}

\newcommand{\baselinectrl}{\pi_0}
\newcommand{\gaitparam}{\beta}
\newcommand{\gaitset}{\mathcal{B}}
\newcommand{\feasgaitset}{\mathcal{B}_{\mathrm{feas}}}
\newcommand{\brt}{BRT}
\newcommand{\statetimegaitsample}{t_i, x_i;\gaitparam_i}
\newcommand{\statetimesample}{t_i, x_i}